\documentclass[10pt,journal,compsoc]{IEEEtran}

\ifCLASSOPTIONcompsoc
  \usepackage[nocompress]{cite}
\else
  \usepackage{cite}
\fi

\ifCLASSINFOpdf
\else
\fi

\usepackage{microtype}
\usepackage{graphicx}
\usepackage{subfigure}
\usepackage{booktabs} 
\usepackage{bm}
\usepackage{amsmath}
\usepackage{arydshln}
\usepackage{multicol}
\usepackage{multirow}
\usepackage{color}

\begin{document}

\title{Controllable Dual Skew Divergence Loss for Neural Machine Translation}

\author{Zuchao Li, Hai Zhao$^*$, Yingting Wu, Fengshun Xiao, and Shu Jiang
	\IEEEcompsocitemizethanks{\IEEEcompsocthanksitem{This paper was partially supported by National Key Research and Development Program of China (No. 2017YFB0304100), Key Projects of National Natural Science Foundation of China (U1836222 and 61733011), Huawei-SJTU long term AI project, Cutting-edge Machine Reading Comprehension and Language Model (Corresponding author: Hai Zhao).}
	\IEEEcompsocthanksitem{Z. Li, H. Zhao, Y. Wu, F. Xiao, and S. Jiang are with the Department of Computer Science and Engineering, Shanghai Jiao Tong University, and also with Key Laboratory of Shanghai Education Commission for Intelligent Interaction and Cognitive Engineering, Shanghai Jiao Tong University, and also with MoE Key Lab of Artificial Intelligence, AI Institute, Shanghai Jiao Tong University. E-mail: charlee@sjtu.edu.cn, zhaohai@cs.sjtu.edu.cn. \protect}
	}
}

\markboth{Arxiv, April~2021}
{Li \MakeLowercase{\textit{et al.}}: Controllable Dual Skew Divergence Loss for Neural Machine Translation}

\IEEEtitleabstractindextext{%
\begin{abstract}
In sequence prediction tasks like neural machine translation, training with cross-entropy loss often leads to models that overgeneralize and plunge into local optima. 
In this paper, we propose an extended loss function called \emph{dual skew divergence} (DSD) that integrates two symmetric terms on KL divergences with a balanced weight. 
We empirically discovered that such a balanced weight plays a crucial role in applying the proposed DSD loss into deep models. 
Thus we eventually develop a controllable DSD loss for general-purpose scenarios.
Our experiments indicate that switching to the DSD loss after the convergence of ML training helps models escape local optima and stimulates stable performance improvements. 
Our evaluations on the WMT 2014 English-German and English-French translation tasks demonstrate that the proposed loss as a general and convenient mean for NMT training indeed brings performance improvement in comparison to strong baselines. 
\end{abstract}

}

\maketitle

\IEEEdisplaynontitleabstractindextext

\IEEEpeerreviewmaketitle

\ifCLASSOPTIONcompsoc
\IEEEraisesectionheading{\section{Introduction}\label{sec:introduction}}
\else
\section{Introduction}
\label{sec:introduction}
\fi

\IEEEPARstart{N}{eural} machine translation (NMT) \cite{kalchbrenner-blunsom,Sutskever14,vaswani2017attention} has shown remarkable performance for diverse language pairs by using the sequence-to-sequence learning framework. 
Unlike the statistical machine translation (SMT) \cite{Koehn2003}, which explicitly models linguistic features of training corpora, NMT aims at building an end-to-end model that directly transforms a source language sequence into the target one \cite{Bahdanau15,luong-pham-manning}. 

During NMT training, maximum likelihood (ML) is the most commonly used strategy, and it maximizes the likelihood of a target sentence conditioned on the source throughout training corpus. 
In practice, ML-based loss is often implemented in a word-level cross entropy form, which has proven to be effective for NMT modeling; however, \cite{RanzatoCAZ16} pointed out that ML training suffers from two drawbacks. 
First, the model is only exposed to training distribution and ignores its own prediction errors during training. 
Second, the model parameters are optimized by a word-level loss during training, while during inference, the model prediction is evaluated using a sequence-level metric such as BLEU \cite{papineni-EtAl:2002:ACL}. 
To mitigate such problems, several recent works focused on the research of more effective and direct training strategies. 
\cite{Bengio2015} advocated a \emph{curriculum learning} approach that gradually forces the model to take its mistakes into account during training predictions as it must do during inference. 
\cite{wiseman-rush} proposed a sequence-level loss function based on errors made during beam search. 
\cite{shen-EtAl} applied minimum risk training (MRT) from SMT to optimize NMT modeling directly in terms of BLEU score. 
Some other works resorted to reinforcement learning based approaches \cite{RanzatoCAZ16,BahdanauBXGLPCB16}.

In this work, we introduce a novel loss called \emph{dual skew divergence} (DSD) to alleviate the issues of the original ML-based loss. 
It can be proved that maximizing the likelihood is equal to minimizing the Kullback-Leibler (KL) divergence \cite{kullback1951} $KL(Q||P)$ between the real data distribution $Q$ and the model prediction $P$. 
According to \cite{Husz2015}, minimizing $KL(Q||P)$ tends to find a $P$ that covers the entire true data distribution and ignores the rest of incorrect candidates, which leads to models that overgeneralize and generate implausible samples during inference. 
As the counterpart of the original $KL(Q||P)$ loss, minimizing the form of $KL(P||Q)$ result in $P$ as the probability mass according to the model's prediction which may correspondingly make the model less trust the data distribution of training data.
To benefit from both of these divergence measures and balance the the overfitting and underfitting in model training, we interpolate $KL(Q||P)$ and $KL(P||Q)$ to form a new DSD loss.
As the two symmetric KL-divergence terms are interpolated by a preset constant (called balanced weight) in the DSD loss, it is shown that such simple setting cannot well handle the case when model training focus may alternate on deep and complicated models. Thus we further propose an adaptive balanced weighting approach by sampling on the divergence, which results in the controllable DSD loss for general model training scenarios.

We carry out experiments on the English-German and English-French translation tasks from WMT 2014 and compare our models to other works with similar model size and matching datasets. 
Our early experiments indicate that switching to DSD loss after the convergence of ML training introduces a simulated annealing like mechanism and helps models quickly escape local optima.
Evaluations on the test sets show that our DSD-extended models both outperform their ML-only counterparts and significantly outperform a series of strong baselines.

\section{Neural Machine Translation}

NMT generally adopts an encoder-decoder model architecture. These architectures roughly fall into three categories: RNN-based \cite{Bahdanau15}, CNN-based \cite{gehring2017convolutional}, and Transformer-based models \cite{vaswani2017attention}.
In our experiments for this work, we include relevant models for these three categories. This section will give a brief introduction to these models.

In an RNN-based NMT model, the encoder is a bidirectional recurrent neural network (RNN) such as one that uses Gated Recurrent Units (GRUs) \cite{cho-EtAl:2014} or a Long Short-Term Memory (LSTM) network \cite{Hochreiter:1997}.
The forward RNN reads an input sequence $x=(x_1, ..., x_m)$ from left to right and calculates a forward sequence of hidden states $(\overrightarrow{h}_1, ..., \overrightarrow{h}_m )$ as the representation of the source sentence. 
Similarly, the backward RNN reads the input sequence in the opposite direction and learns a backward sequence $(\overleftarrow{h}_1, ..., \overleftarrow{h}_m)$. The hidden states of the two RNNs $\overrightarrow{h}_i$ and $\overleftarrow{h}_i$ are concatenated to obtain the source annotation vector $h_i=[\overrightarrow{h}_i, \overleftarrow{h}_i]^T$ as the initial state of the decoder.

The decoder is a forward RNN that predicts a corresponding translation $y = (y_1, ..., y_n)$ step by step. The translation probability can be formulated as follows:
\begin{equation}\nonumber
p(y_j|y_{<j}, x) = q(y_{j-1}, s_j, c_j), 
\end{equation}
where $s_j$ and $c_j$ denote the decoding state and the source context at the $j$-th time step, respectively. Here, $q(\cdot)$ is the softmax layer and $y_{<j}=(y_1, ..., y_{j-1})$. Specifically, 
\begin{equation}\nonumber
s_j = g(y_{j-1}, s_{j-1}, c_j),
\end{equation}
where $g(\cdot)$ is the corresponding RNN unit. The context vector $c_j$ is calculated as a weighted sum of the source annotations $h_i$ using the attention mechanism:
\begin{equation}\nonumber
c_j = \sum_{i=1}^{m}\alpha_{ji}h_i.
\end{equation}
The alignment model $\alpha_{ji}$, a single layer feed-forward neural network, gives the probability of how well $y_j$ is aligned to $x_i$.
The whole model is jointly trained to seek the optimal parameters to sufficiently encode the source sentences and decode them to their corresponding target sentences.

In a CNN-based NMT model, instead of relying on RNNs to compute intermediate encoder states $h_i$ and decoder states $s_j$, Convolutional Neural Networks (CNNs) are adopted, which makes the model a fully convolutional architecture for the sequence to sequence modeling. 
To preserve position information in the CNN encoding, position embedding is also included in the model.

Transformer-based NMT leverages multi-headed self-attention to replace RNN and CNN in the encoder's and decoder's respective $h_i$ and $s_j$ states. By leveraging attention mechanisms, the model can disperse recurrence and convolutions completely. Similarly, position embedding should also be adopted in order to prevent the loss of position in encodings.

\section{Dual Skew Divergence Loss for NMT}

\subsection{Cross Entropy}

In information theory, cross entropy is an important concept that measures the difference between two probability distributions. 
In natural language processing, cross entropy is usually used as the evaluation metric for language modeling. 
During NMT training, word-level cross entropy broadly serves as the loss function for learning the model parameters.

NMT aims at training a model that directly transforms a given source sequence $x=(x_1, x_2, ..., x_m)$ to its corresponding target sequence $y=(y_1, y_2, ..., y_n)$. 
Given a set of training examples $D=\{\langle x^{(k)}, y^{(k)}\rangle\}_{k=1}^N$, the standard ML training objective is to maximize the log-likelihood of the training data with respect to the parameters $\bm{\theta}$:
\begin{equation*}
\begin{split}
\hat {\bm{\theta}}_\mathrm{ML} &= \mathop{\mathbf{argmax}}_{\bm{\theta}} \{\mathcal L (\bm{\theta})\},\\
\mathcal L (\bm{\theta}) &= \sum_{k=1}^N\log P(y^{(k)}|x;\bm{\theta}) \\
&= \sum_{k=1}^N\sum_{i=1}^{n}\log P(y_i^{(k)}|x^{(k)}, y_{<i}^{(k)};\bm{\theta}). \\
\end{split}
\end{equation*}

To accommodate the gradient descent method, this objective is actually transformed to minimizing the negative log-likelihood. In practice, this is realized by minimizing the word-level cross entropy, which can be calculated at each time step of decoding. Given an observed target sequence of length $n$, the cross entropy loss in the vector form can be represented as
\begin{equation*}\label{eq1}
J_\mathrm{XENT}=-\sum_{i=1}^{n}\mathbf{y_i}\log(\mathbf{\hat{y}_i}),
\end{equation*}
where $\mathbf{y_i}$ is a one-hot vector; referring to the correct target label at time step $i$, and $\mathbf{\hat{y}_i}$ is the model approximation given by the softmax layer.

\subsection{Kullback-Leibler Divergence}

Kullback-Leibler (KL) divergence \cite{kullback1951} is also a measurement that calculates the distance between two probability distributions and is denoted by: 
\begin{equation}\label{eq3}
D_{KL}(Q||P)=E_{x\sim Q}[\log Q(x)-\log P(x)],
\end{equation}
where $Q$ is the true data distribution of target words, and $P$ is the distribution of model prediction.

Mathematically, there is a connection between KL divergence $D_{KL}(Q||P)$ and cross entropy $H(Q, P)$ that can be described as follows:
\begin{equation*}
H(Q, P) = H(Q) + D_{KL}(Q||P),
\end{equation*}
where $H(Q)$ refers to the entropy of the training data itself. Since the gradient of $H(Q)$ with respect to the model parameters that describe $P$ is always equal to zero, training with $H(Q, P)$ is identical to that with $D_{KL}(Q||P)$. In other words, minimizing cross entropy loss actually means minimizing KL divergence.

\subsection{Asymmetry of KL divergence}

Swapping the positions of $P$ and $Q$, we can easily get $D_{KL}(P||Q)$, divergence in the inverse direction of $D_{KL}(Q||P)$. Following Eq.(\ref{eq3}), $D_{KL}(P||Q)$ can be also rewritten as 
\begin{equation*}
D_{KL}(P||Q)=E_{x\sim P}[-\log Q(x)+\log P(x)].
\end{equation*}
Obviously, $D_{KL}(Q||P)$ is not identical to its inverse form $D_{KL}(P||Q)$. 
$D_{KL}(Q||P)$ and $D_{KL}(P||Q)$ on $N$ training samples can be written as:
\begin{equation*}
\begin{split}
D_{KL}(Q||P) &= \sum_{i=1}^{n} Q(x_i) \log \frac{Q(x_i)}{P(x_i)},\\
D_{KL}(P||Q) &= \sum_{i=n}^{N} P(x_i) \log \frac{P(x_i)}{Q(x_i)}.
\end{split}
\end{equation*}
In minimizing the $D_{KL}(Q||P)$ divergence, since $Q$ is a true data distribution, it is one-hot when without smoothing, that is, the correct place is 1 and the wrong place is 0. 
The meaning is to drive the model's prediction distributed similarly in the correct place while ignoring all the wrong places.
$Q(x)$ in $D_{KL}(Q||P)$ as a probability mass is determined by the true data distribution in $D_{KL}(Q||P)$ formula.
In other words, the model's focus on the similarity of the distribution is determined by the probability quality.
While in minimizing the $D_{KL}(Q||P)$ divergence, the model's focus on the distribution similarity is determined by the distribution of model's  prediction $P(x)$ instead of the true data distribution.

In the training of NMT model, the model is essentially required to predict a high probability on the correct word as much as possible, which can be done through $D_{KL}(Q||P)$. 
Meanwhile, there are some facts that need to be considered in NMT. 
First, the presence of noise in the training data samples leads to inaccuracy of $Q$ and the existence of multiple target candidates in translation leads to inadequate of $Q$. 
Overemphasizing the distribution similarity of the correct position will lead to overfitting or poor generalization ability. 
Second, if we completely rely on the predictive distribution as the probability mass in $D_{KL}(P||Q)$, it is likely that the model cannot converge, because the model cannot make a correct estimate of the data distribution in the early training period.
In general, the two different objectives have their own advantages and disadvantages, so it is hard to say which is actually better. 
Despite this, their differences motivate us to find a more proper loss function.

\subsection{Skew Divergence}

The above analysis shows that $D_{KL}(P||Q)$ may be also a good alternative loss function for NMT training. However, $D_{KL}(P||Q)$ is only well-defined when $Q$ is not equal to zero, which is not guaranteed in NMT training. 
To overcome the this, we will use two different approximation functions instead. One such approximation function is the $\alpha$-\emph{skew divergence} \cite{lee:1999:ACL}:
\begin{equation}\nonumber
s_\alpha(P, Q) = D_{KL}(P||\alpha P+(1-\alpha)Q),
\end{equation}
where $\alpha$ controls the degree to which the function approximates $D_{KL}(P||Q)$, and $0\leq \alpha \leq 1$. 

By assigning a small value to $\alpha$, we can simulate the behavior of minimizing $D_{KL}(P||Q)$ with an $\alpha$-\emph{skew divergence} approximation. In our experiments, $\alpha$ is set to to a constant 0.01.

\begin{figure*}[!htbp]
	\centering 
	\includegraphics[width=0.97\textwidth]{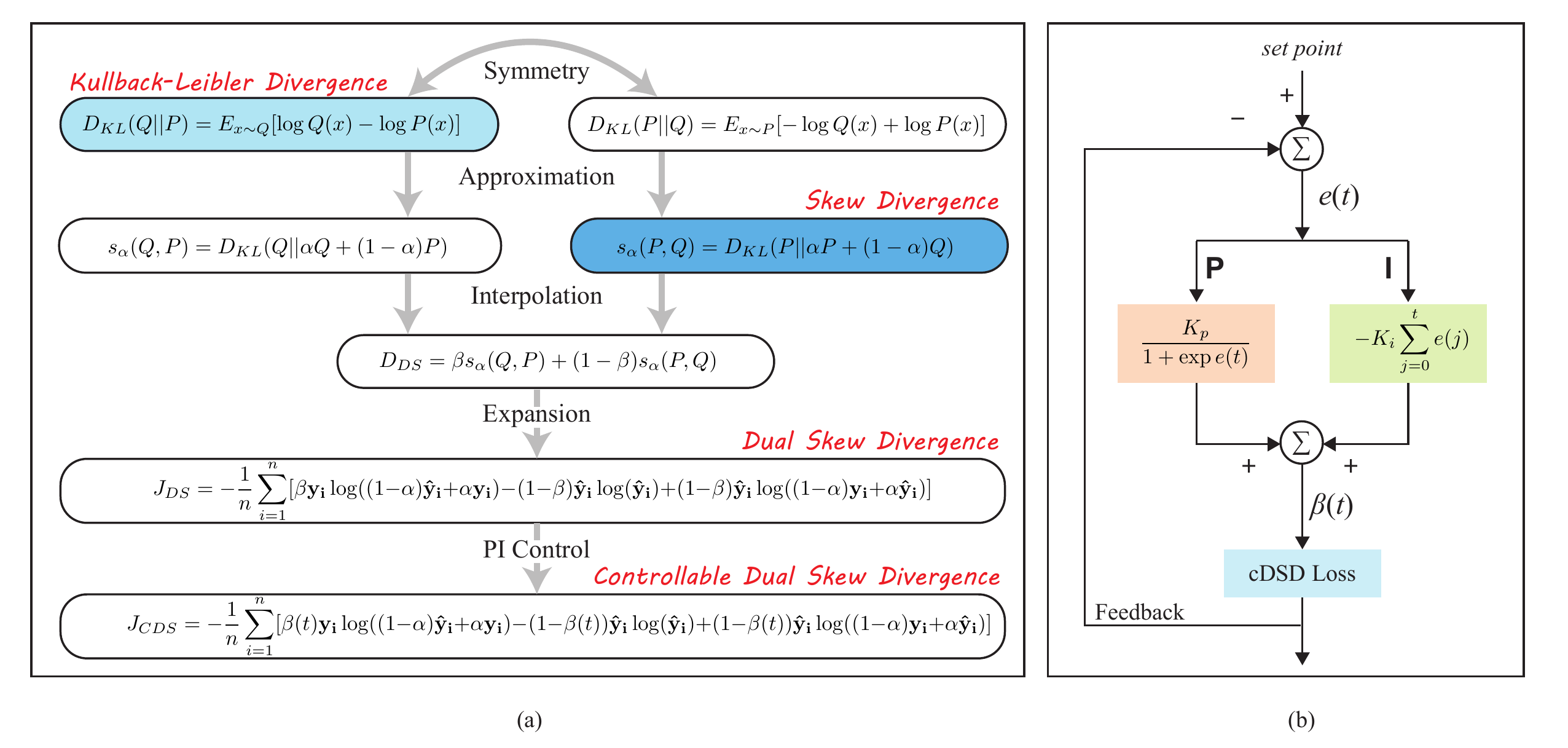} 
	\caption{(a) The derivation of our proposed DSD and cDSD loss function. (b) PI controller in cDSD loss.}\label{illus}
\end{figure*}


\subsection{Dual Skew Divergence}\label{sec_DSD}

Finally, we introduce a new loss with the aim of retaining the strengths of the ML objective while minimizing its weaknesses. 
In order to obtain a symmetrical form of the loss, we also approximate $D_{KL}(Q||P)$ with $s_\alpha(Q, P)$. By interpolating between both directions of $\alpha$-\emph{skew divergence}, we derive the new loss function, which we refer to as \emph{dual skew divergence} (DSD), below. The interpolated DSD function is given by 
\begin{equation} \label{eq2}
D_{DS}=\beta s_\alpha(Q, P)+ (1-\beta) s_\alpha(P, Q),
\end{equation}
in which $\beta$ is called balanced weight as the interpolation coefficient, and $0\leq \beta \leq 1$. 

In terms of notations in Eq.(\ref{eq1}), we rewrite Eq.(\ref{eq2}) in a computationally implementable form as below:
\begin{eqnarray*}
	J_{DS}=-\frac{1}{n}\sum_{i=1}^{n}[\beta \mathbf{y_i}\log((1-\alpha)\mathbf{\hat{y}_i}+\alpha \mathbf{y_i}) \\
	-(1-\beta) \mathbf{\hat{y}_i}\log(\mathbf{\hat{y}_i}) \\
	+(1-\beta) \mathbf{\hat{y}_i}\log((1-\alpha)\mathbf{y_i}+\alpha \mathbf{\hat{y}_i})],
\end{eqnarray*}
where $n$ is the length of target sequence, and the term about $H(Q)$ is omitted since it does not affect the gradient as discussed above. 
Note that in addition to the above equation, we also add a small constant ($10^{-12}$ in our implementation) to each log term for numerical stability. 
Our derivation of the proposed DSD loss is summarized in the step by step diagram of Figure \ref{illus}.

\textcolor{black}{In machine translation, the DSD loss is ideally used to reduce overfitting of the model from the target in the training set and robustness losing as compared to word-level ML training. In the mainstream ML training in machine translation, a smoothing mechanism \cite{chen-goodman-1996-empirical} is usually adopted for similar purposes. There is an inconsistency with the reality of machine translation since ML+smoothing actually gives all incorrect tokens an equal probability of being correct due to special phenomena like synonyms and synonyms. Assigning equal probability to incorrect tokens is also referred to as \textit{negative diversity ignorance} problem, which is proposed in \cite{Li2020Data-dependent}. Compared with the D2GPo loss in \cite{Li2020Data-dependent}, although the motivation is similar, the D2GPo loss mainly relies on the prior knowledge from the pre-training embedding for incorrect tokens, which provide a more flexible and realistic distribution; while the DSD loss does not rely on the external distribution at all, it only relies on the model's own estimation of the predicted token distribution. In DSD loss, symmetrical loss calculations are built, one is optimized based on the correct token, and the other is optimized based on the model's own estimation. Through adversarial-like training between loss items, the ultimate goal is achieved.}

\subsection{Controllable Dual Skew Divergence}

Our proposed dual skew divergence consists of two terms: symmetric skew divergence $s_{\alpha}(Q, P)$ and $s_{\alpha}(P, Q)$. 
The first term is intended to focus the model on the correct word only, while the second term uses the focus distribution, estimated by itself, to learn the distribution of all words.
These two influence each other and sometimes conflict.
When $s_{\alpha}(Q, P)$ is weighted more heavily, the model bases its predictions more on the true distribution $Q$ of the training data, and when $s_{\alpha}(P, Q)$ is weighted more heavily, the model assigns the weight of the loss of each word based on its own prediction distribution. This conflict and influence between the two divergences makes the training somewhat adversarial. 
To balance the two terms, the balanced weight $\beta$ is introduced to control the training process.

Since the training is dynamic, a fixed $\beta$ in the dual skew divergence loss cannot effectively adapt to the training for deep and hyperparameter-sensitive CNN-based and Transformer-based models. 
The key challenge in applying DSD loss to deeper and hyperparameter-sensitive models lies in the difficulty of tuning the $\beta$ of the divergence term during model training. 
Inspired by control systems and ControlVAE \cite{shao2020controlvae}, we further propose a controllable dual skew divergence (cDSD) loss with a feedback control to address this inconvenience in the original DSD.

Specifically, in the cDSD loss, we replaced the balance weight $\beta$ with a time step related version $\beta(t)$.
During training, we sample the divergence term $s_{\alpha}(Q, P)$ at each training step $t$. We denote this sample as $u(t)$; and tune $\beta(t)$ with a controller to stabilize the divergence at a desired value $u^*$, called the \textit{set point}. The cDSD loss becomes:
\begin{eqnarray*}
	J_{CDS}=-\frac{1}{n}\sum_{i=1}^{n}[\beta(t) \mathbf{y_i}\log((1-\alpha)\mathbf{\hat{y}_i}+\alpha \mathbf{y_i}) \\
	-(1-\beta(t)) \mathbf{\hat{y}_i}\log(\mathbf{\hat{y}_i}) \\
	+(1-\beta(t)) \mathbf{\hat{y}_i}\log((1-\alpha)\mathbf{y_i}+\alpha \mathbf{\hat{y}_i})],
\end{eqnarray*}

PID control \cite{aastrom2006advanced} is a basic (and the most prevalent) form of the feedback control algorithm in a wide range of industrial and software performance controls. The general form of the PID controller is defined as:
\begin{equation*}\label{eq:pid}
\beta(t) = K_p e(t) + K_i \int_0^t e(\tau)d\tau + K_d \frac{de(t)}{dt},
\end{equation*}
in which an error $e(t)$ between the \textit{set point} and the value in time step $t$ is calculated to potentially trigger a correction and reduce the error through the output of the controller $\beta(t)$; $K_p, K_i$, and $K_d$ are the coefficients for the P term (changes with the  error), I term (changes with the integral of the error) and D term (changes with the derivative of the error), respectively.

Following existing practices in the control of noisy systems, we adopt a variant of the PID controller: a nonlinear PI controller for $\beta(t)$ tuning, which eliminates the use of the derivative (D) term in our controller. Our controller can be expressed as follows:
\begin{equation}\nonumber
\beta(t) = \min(\beta_{max}, \frac{K_p}{1+\exp(e(t))} - K_i \sum_{j=0}^t e(j) + \beta_{min}),
\end{equation}
where $K_p$ and $K_i$ are both constants.
With $\beta(t)$, when an error is large and positive, $s_{\alpha}(Q, P)$ is below the \textit{set point}; that is, the likelihood loss of the model on the training set is very small, and the model is very likely to overfit. Thus, the first term in controller approaches 0, leading to a lower $\beta(t)$ that encourages the model focus on the word estimated by the model itself instead of the distribution given by the training data . %
When the error is large and negative, the first term approaches its maximum value (i.e. $K_p$) and the resultant higher $\beta(t)$ value leads to the model focusing more on the correct word given by the training data.
For the second term, errors within $T$ training steps (we use $T=1$ in this work) are summed to create a progressively stronger correction. $\beta_{min}$ and $\beta_{max}$ are application-specific constants that effectively constrain the range within which $\beta(t)$ can vary.

As discussed in \cite{shao2020controlvae}, constants $K_p$ and $K_i$ in the PI controller should guarantee that error reactions are sufficiently smooth to allow for gradual convergence. Following the settings in ControlVAE \cite{shao2020controlvae}, in this work, we also let $K_p=0.01$ and $K_i=0.0001$.
For the \textit{set point} of $s_{\alpha}(Q, P)$, its upper bound is the divergence value when the model converges with fixed $\beta = \beta_{min}$ in the original DSD loss. Similarly, its lower bound is the divergence value when $\beta = \beta_{max}$ in the original DSD loss. Since the value of \textit{set point} is highly customizable to satisfy its different applications, we use the same divergence value that ML's cross entropy loss with label smoothing calculates as the \textit{set point} in our empirical study.

\textcolor{black}{To conform to the training scenarios of deeper and parameter-sensitive models, we propose the cDSD loss. Although model training with DSD loss is not adversarial, the two KL terms in DSD play an adversarial role for the model in essential. Similar to the adversarial training in a GAN \cite{goodfellow2014generative} model, it may be challenging to train a stable model, especially when there are many parameters that need to be optimized. The reason is that the training process is inherently unstable, resulting in the simultaneous dynamic training of two competing items (models/objectives). In other words, it is a dynamic system in which the optimization process seeks an equilibrium between two forces rather than a minimum. Since the skew divergence in DSD is a term that introduces smoothing or noise to the target, when we keep this noise term at a close to a constant value, the model can run in a stable manner. PI control is to control the loss of smoothing within a reasonable range, which is why cDSD loss is more suitable for deeper models. In the DSD loss, the fixed $\beta$ makes it impossible to adjust according to the changes in the training of the model; an inappropriate smoothing loss value will make the training unstable.}

\section{Experiments}

\subsection{Data Preparation}

We perform all experiments on data from the shared task of WMT 2014 and report results on the English-German and English-French translation tasks. 
The translation quality is measured by case-sensitive 4-gram BLEU score \cite{papineni-EtAl:2002:ACL}, and we use the sign test \cite{collins2005clause} to test the statistical significance of our results.

For the English-German task, the training set consists of 4.5M sentence pairs with 91M English words and 87M German words. 
For the English-French task, the training set contains 39M sentence pairs with 988M English words and 1131M French words. The models are evaluated on the WMT 2014 test set \emph{news-test} 2014, and the concatenation of \emph{news-test} 2012 and \emph{news-test} 2013 is used as the development set. 

The preprocessing on both training sets includes a joint byte pair encoding \cite{sennrich-haddow-birch} with 32K merge operations after tokenization. The final joint vocabulary sizes are around 37K and 37.2K for the English-German and English-French translation tasks, respectively. Every out-of-vocabulary word is replaced with a special $\langle$UNK$\rangle$ token. 

\subsection{Models and Training Details}

We train both a baseline SMT system and serveral NMT baseline systems.
For the SMT baseline, we use the phrase-based SMT system M{\scriptsize{OSES}} \cite{Koehn2007Moses}. 
The log-linear model of M{\scriptsize{OSES}} is trained by minimum error rate training (MERT) \cite{och:2003:ACL}, which directly optimizes model parameters with respect to evaluation metrics. 
Our SMT baseline is trained with the default configurations in M{\scriptsize{OSES}}, and it is trained together with a trigram language model trained on the target language using SRILM \cite{Stolcke2002Srilm}. 
For the RNN-based NMT baseline, we use the model architecture of the attention-based RNNSearch \cite{Bahdanau15}. 
Our RNN-based NMT baseline model is generally similar to \cite{Bahdanau15}, except we apply an input feeding approach, and the attention layer is built on top of a LSTM layer instead of a GRU layer.
For the CNN-based NMT baseline, we use the same architecture of model ConvS2S which is introduced in \cite{gehring2017convolutional}. For the Transformer-based NMT baseline, we adopt the Transformer model presented in \cite{vaswani2017attention} without other model structure changes.

As in our RNN-based baseline model, each direction of the LSTM encoder and the LSTM decoder has dimension 1000.
The word embedding and attention sizes are both set to 620. 
The batch size is set to 128, and no dropout is used for any model. 
The CNN-based baseline has 20 convolutional layers in both the encoder and the decoder, and both use kernels of width 3 and hidden size 512 throughout.
In the Transformer-based model, there are two commonly used parameter settings: Transformer-\textit{base} and Transformer-\textit{big}. 
We choose the best performing Transformer-\textit{big} for experiments, consisting of 6 layers each for the encoder and decoder with a hidden size of 1024, 16 attention heads, and 4096-dimensional feed-forward inner-layers.

The models trained with our DSD loss and cDSD loss are referred to DSD-NMT and cDSD-NMT, respectively, hereafter.
To apply the proposed DSD loss, we adopt a hybrid training strategy. 
To provide a reliable initialization, we start training the model with cross entropy loss and then switch to the DSD loss at different switching points. 
The training set is reshuffled at the beginning of each epoch. 8 Nvidia Tesla V100 GPUs are used to train all the NMT models.
For the English-German task, training lasts for 9 epochs in total. 
We use the Adam optimizer for the first 5 epochs with a learning rate of $3.0 \times 10^{-4}$; and then switch to plain SGD with a learning rate of 0.1. 
At the beginning of epoch 8, we decay the learning rate to 0.05. 
For the English-French task, the models are trained for 4 epochs. 
The Adam optimizer and a learning rate of $3.0 \times 10^{-4}$ are used for the first 2 epochs. 
We then switch to SGD with a learning rate of 0.1; and finally decay the learning rate to 0.05 at the beginning of epoch 4.
In the training of cDSD-NMT, $beta(0)$ is initialized to 1, and the \textit{set point} for divergence $s_{\alpha}(Q, P)$ is set to 35 and 33 for the English-German and English-French tasks, respectively. $\beta_{min}$ is set to 0.85, and $\beta_{max}$ is set to 0.95 for both tasks.

To demonstrate the source of improvement with our approach, we mainly use the RNN-based model as the basis of analysis, we therefore compare our DSD-NMT model to several important RNN-based NMT systems with the same dataset and similar model size.
\begin{itemize}
	\item RNNSearch-LV \cite{chousing}: a modified version of RNNSearch based on importance sampling, which allows the model to have a very large target vocabulary without any substantial increase in computational cost.
	\item Local-Attn \cite{luong-pham-manning}: a system that applied a local attention mechanism that focuses only on a small subset of the source positions when predicting each target word.
	\item MRT \cite{shen-EtAl}: a system optimized by a loss function for minimum risk training. The model parameters are directly optimized with respect to the evaluation metrics.
	\item Bahdanau-LL \cite{BahdanauBXGLPCB16}: this model closely followed the architecture of \cite{Bahdanau15} with ML training and achieved a higher performance by annealing the learning rate and penalizing insufficiently long output sequences during beam search.  
	\item Bahdanau-AC+LL \cite{BahdanauBXGLPCB16}: a neural sequence prediction model that combines the actor-critic from reinforcement learning with the original ML training.
\end{itemize}

\subsection{Effect of Balanced Weight $\beta$}

As shown in Section \ref{sec_DSD}, the balanced weight $\beta$ controls the degree of conservativeness of the model. 
When $\beta$ is close to one, the loss function behaves more like $D_{KL}(Q||P)$, which tends to predict the correct target on the training dataset as accurate as possible.
When $\beta$ is close to zero, the loss function behaves more like $D_{KL}(P||Q)$, which prefers to conservatively ignore words that are considered correct in the training set and to choose correct words that match the model's predictions.
In RNN-based DSD-NMT, in order to find an optimal value of $\beta$, we study the effect of $\beta$ on the translation quality of the English-German task.
In addition, we also include the results of RNN-based cDSD-NMT for better analysis.
Table \ref{tab1} reports the BLEU scores with different $\beta$ in RNN-based DSD-NMT and cDSD-NMT on the development set with greedy search.
When only taking the DSD-NMT results into consideration, the results seem to show that the model trained with $\beta=0$ performs the best; and suggest that the skew inverse KL divergence outperforms the interpolation form in our loss function-applying strategy, but when comparing DSD-NMT with that with cDSD-NMT, we see that cDSD-NMT obtains slightly better results with the same model.
On the one hand, this shows that dynamic $\beta$ adjustment has advantages over the fixed $\beta$ setting, and on the other hand, it also shows that the interpolation form is actually a more general loss than a single divergence.
\textcolor{black}{It is worth noting that when $\beta=1$, optimizing DSD is actually equivalent to MLE loss, and a performance improvement of 0.4 is still observed compared to the baseline. The source of these improvements may be attributed to the Adam-SGD switching training method. This switching of optimizer and learning rate has a possibility of breaking optimization from a local optimum.}

\begin{table}[!htbp]
	\centering
	\small
	\caption{BLEU scores with different $\beta$ in RNN-based DSD-NMT and cDSD-NMT on English-German dev set.}\label{tab1}
	\begin{tabular}{cccccc}
		\toprule
		\multirow{2}{*}{\textbf{Model}} & \multirow{2}{*}{\textbf{Baseline}} & \multicolumn{3}{c}{\bf DSD} & \multirow{2}{*}{\textbf{cDSD}}\\
		& & $\beta=1$ & $\beta=0.5$ & $\beta=0$ & \\
		\midrule
		\textbf{BLEU} & 20.51 & 20.91 & 21.34 & \textbf{21.72} & \textbf{21.96} \\
		\bottomrule
	\end{tabular}
\end{table}

\begin{figure}[!htbp]
	\centering
	\includegraphics[width=1.0\linewidth]{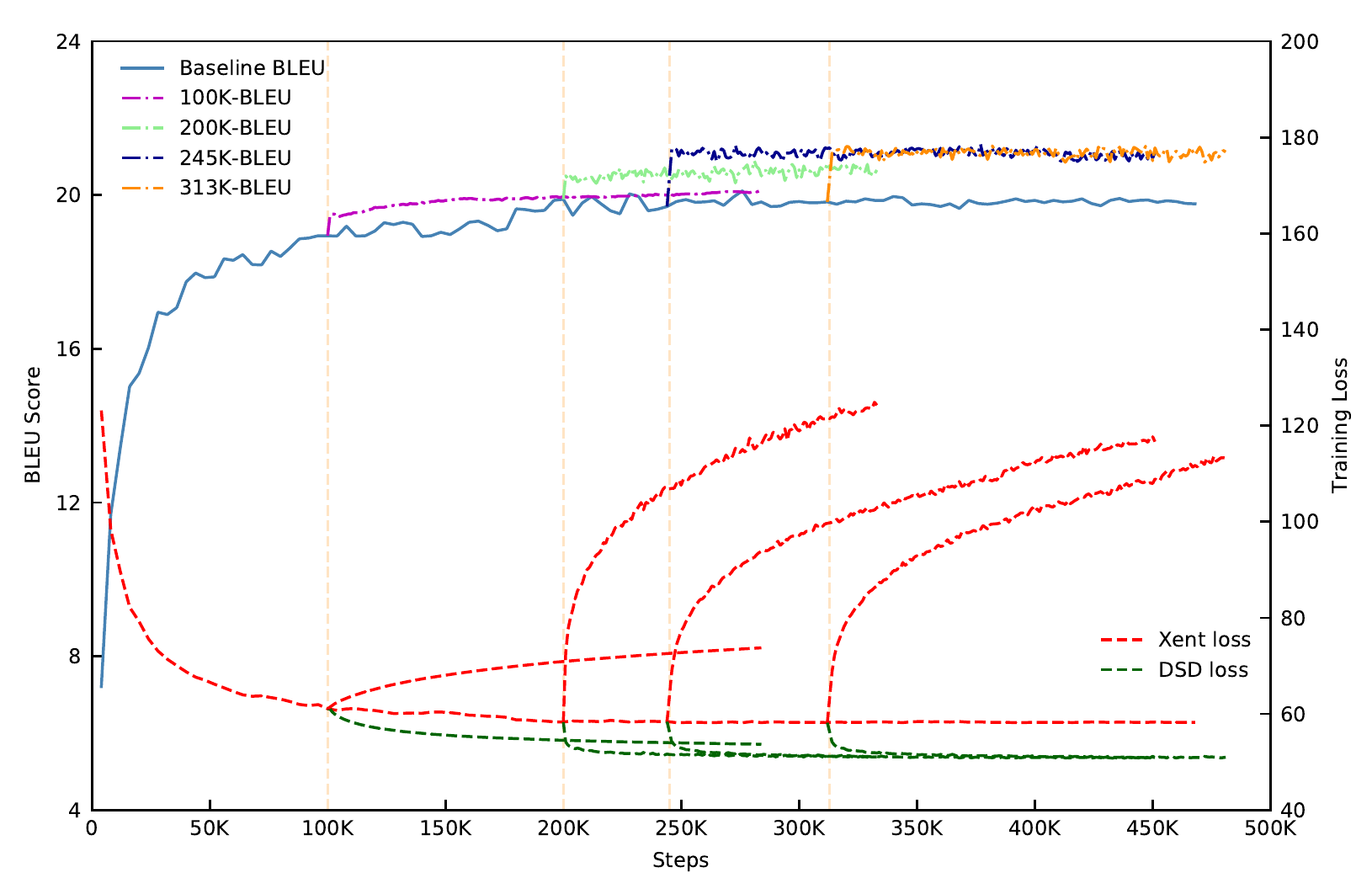} 
	\caption{BLEU scores and training loss on
		English-German dev set with DSD switching after 100K, 200K, 245K and 313K steps ($\beta = 0$, greedy search). }\label{fig1} 
\end{figure}

\subsection{Switching Point and BLEU}
\textcolor{black}{Since the DSD loss contains two adversarial loss items, if DSD is leveraged in the early training stage when the model has not yet converged yet, the model may oscillate at a lower performance level, which will affect the convergence speed and final convergence performance. Therefore, we first use the ML training model to achieve rapid convergence and then switch to DSD loss for better performance finetuning.}
Since we start the training of DSD-NMT and cDSD-NMT with an initial ML-trained model, the model performance with this training strategy will be influenced by when the model switches to the new loss function.
To intuitively show the relationship between the switching point and BLEU, we plot the curves of BLEU score and training loss against training steps from the actual training process of RNN-based DSD-NMT for the English-German task with greedy search at different switching positions (after 100K, 200K, 245K, and 313K steps) in Figure \ref{fig1}, . 
From the figure, it shows that when switching to our new loss function, the BLEU scores of both settings are improved. 
In particular, there is a more than one point improvement at steps 245K and 313K. 
Comparing all the training curves shows that a better DSD switching point should be located around the convergence of the standard ML training. 
After switching to DSD, the cross entropy loss (which should theoretically be minimized) actually increases, as do the BLEU scores, which demonstrates that cross entropy fails to reflect translation quality near the end of training. 
Sufficiently long time training may cause the model to plunge into a local optimum, but the loss switching operation resembles a simulated annealing mechanism\footnote{We also tried an automatically loss function switching strategy similar to simulated annealing by switching the loss according to the growth rate of BLEU; however, switching DSD back to cross entropy does not further increase BLEU over the original score.} and helps the model move to a better optimum.

\subsection{DSD and cDSD in CNN-based and Transformer-based Models}

\begin{table}[]
    \centering
	\small
	\caption{BLEU scores with different model architectures and loss functions on the English-German dev set.}
    \label{tab:loss}
    \begin{tabular}{c|ccc}
         \toprule
         & \bf RNN-based & \bf CNN-based & \bf Transformer-based \\
         \midrule
         Baseline & 20.11 & 22.64 & 24.76\\
         DSD & 21.32 & 22.60 & 24.55 \\ 
         cDSD & 21.44 & 22.98 & 25.13 \\
         \bottomrule
    \end{tabular}
\end{table}

In order to illustrate the effects of DSD and cDSD on different baselines, we conducted experiments on English-German translation; and performed general ML training, DSD training, and cDSD training using the three baselines. The results are shown in Table \ref{tab:loss}.
From the results, first, cDSD is generally stronger than DSD in all three model architectures. 
In RNN-based NMT, the gap between cDSD and DSD is relatively small. Therefore, in order to simplify training, we only use DSD on the RNN-based model in subsequent experiments. 
On the CNN-based and Transformer-based NMT models, the results of using DSD are even inferior to the baseline, but the cDSD continues to bring improvements. 
We speculate that the reason is that as the models deepened, the structure became more complex, making the training process more sensitive, so the static $\beta$ in DSD loss was no longer applicable. Therefore, we used the cDSD loss for training the CNN and Transformer models.

\begin{table}
		\centering
		\small
		\caption{Performance on the WMT14 English-German task}\label{tab2}
		\begin{tabular}{lll}
		\toprule
			\textbf{Model} & \textbf{Method} & \textbf{BLEU} \\
			\midrule
			RNNsearch-LV & ML+beam & 19.40 \\
			Local-Attn & ML+beam & 20.90 \\
			MRT & MRT+beam & 20.45 \\
			Baseline-SMT & MERT+greedy & 18.83 \\
			& MERT+beam  & 19.91 \\
			\hdashline
			\multirow{6}{*}{RNN-based NMT} & ML+greedy  & 20.89 \\
			& ML+beam  & 22.13 \\
			& ML+deep  & 24.64 \\
			& \bf DSD+greedy & $\textbf{22.02}^{++}$ \\            
			& \bf DSD+beam  & $\textbf{22.60}^{+}$ \\
			& \bf DSD+deep & $\textbf{25.00}^{++}$ \\ 
			\midrule
			CNN-based NMT & ML+beam & 26.43 \\
			& \bf cDSD+beam & $\textbf{26.72}^{++}$ \\
			\midrule
			Transformer-based NMT & ML+beam & 28.32 \\
			& \bf cDSD+beam & $\textbf{28.64}^{+}$ \\
			\bottomrule
		\end{tabular}
\end{table}

\begin{table}
	\centering
	\small
	\caption{Performance on the WMT14 English-French task}\label{tab3} 
	\begin{tabular}{lll}
	    \toprule
		\textbf{Model} & \textbf{Method} & \textbf{BLEU} \\
		\midrule
		RNNsearch-LV & ML+beam  & 34.60 \\
		MRT & MRT+beam  & 34.23 \\
		Bahdanau-LL & ML+greedy  & 29.33 \\
		& ML+beam  & 30.71 \\
		Bahdanau-AC+LL & ML+AC+greedy  & 30.85 \\
		& ML+AC+beam  & 31.13 \\
		\midrule
		Baseline-SMT & MERT+greedy  & 31.55 \\
		& MERT+beam  & 33.82 \\
		\hdashline
		\multirow{4}{*}{RNN-based NMT} & ML+greedy  & 32.10 \\
		& ML+beam  & 34.70 \\
		& \bf DSD+greedy  & $\textbf{33.56}^{++}$ \\
		& \bf DSD+beam & $\textbf{35.04}^{+}$ \\
		\midrule
		CNN-based NMT & ML+beam & 41.44 \\
		& \bf cDSD+beam & $\textbf{41.72}^{+}$ \\
		\midrule
		Transformer-based NMT & ML+beam & 41.79 \\
		& \bf cDSD+beam & $\textbf{42.00}^{+}$ \\
		\bottomrule
	\end{tabular}
\end{table}

\subsection{Results on Test Sets}

The results\footnote{``++'' indicates a statistically significant difference from the NMT baseline at $p < 0.01$ and ``+'' at $p  <0.05$.} on the English-German and English-French translation test sets are reported in Tables \ref{tab2} and \ref{tab3}. 
For previous works, the best BLEU scores of single models from the original papers are listed. 
From Tables \ref{tab2} and \ref{tab3}, we see that RNN-based DSD-NMT model outperforms all the other models and our own baselines that use standard cross entropy loss with greedy search or beam search.
For the English-German task, Table \ref{tab2} shows that even our RNN-based NMT baseline models achieve better performance than most of the listed systems, though this result may be due to the use of joint BPE, input feeding, and the mixed training strategy using the Adam and SGD algorithms. 
When using greedy search, our DSD model outperforms the SMT and NMT baselines by 3.19 and 1.13 BLEU , respectively. 
This is even better than the best listed system, Local-Attn \cite{luong-pham-manning}, which uses beam search. 
When using beam search, RNN-based DSD-NMT outperforms the SMT baseline with improvement of 2.69 BLEU points; however, it only provides a 0.47 BLEU increase over the NMT baseline. 
We also test our DSD loss on a deep NMT model where the encoder and decoder are both stacked 4-layer LSTMs. 
The result indicates a comparatively less 0.36 point gain, which illustrates both the usefulness of our DSD loss and that the improvement decreases as the depth increases.

For the English-French task, with greedy search, the performance of DSD-NMT is still superior to other systems listed in Table \ref{tab3}. It achieves an increase of 1.46 and 2.01 BLEU points compared to the NMT and SMT baselines, respectively. 
With beam search, our DSD-NMT outperforms the SMT and NMT baselines by 1.22 and 0.34 BLEU, respectively. 
Both tables' sigh test results also indicate that DSD-NMT indeed significantly enhances the translation quality in comparison baselines trained with only cross entropy loss. 

The CNN-based and Transformer-based NMT models, whose network depth reaches 20 and 12 layers, respectively, have stronger performance than the RNN-based NMT model. 
Additionally, on these strong baselines, our cDSD loss still brings a stable improvement in translation performance for both English-German and English-French, indicating that the DSD we proposed is a general loss for NMT model training.

\section{Ablation Study}

\subsection{\textcolor{black}{Greedy and Beam Search}}

\begin{figure}[!htbp] 
	\centering 
	\includegraphics[width=1.0\linewidth]{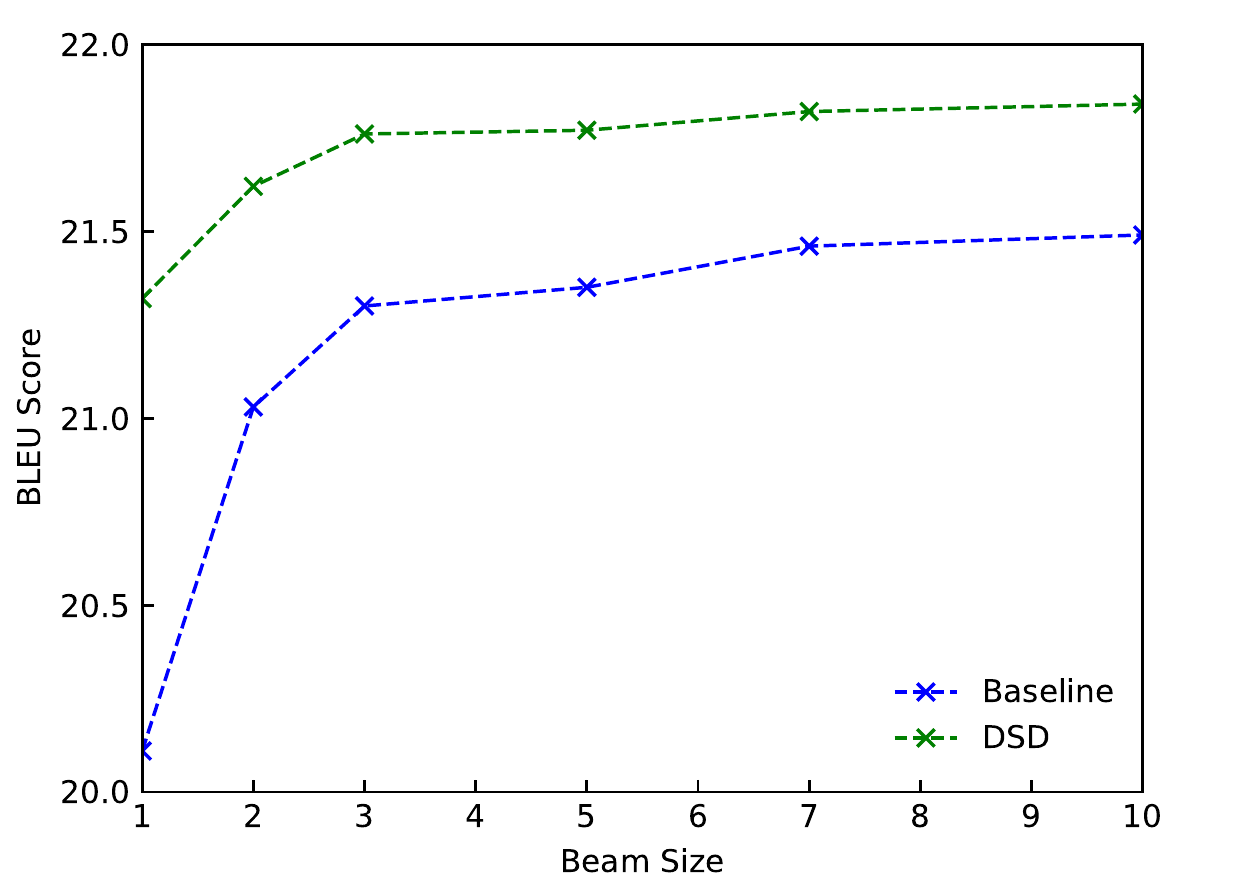} 
	\caption{BLEU scores with different beam sizes
		on English-German dev set($\beta = 0.5$).}\label{fig2} 
\end{figure}

\textcolor{black}{Beam search is a commonly used method to jump out of the local optimal in the inference stage. In order to compare the synergy between beam search and DSD loss during training, } we plot the variation curves of BLEU scores for the proposed DSD model and the NMT baseline with different beam sizes on the English-German development set in Figure \ref{fig2}. 
With increasing beam size, the BLEU score also increases, and the best score is given by the beam size of 10 for both the NMT baseline and DSD-NMT. 
When beam search is used, however, the margin between the proposed DSD method and ML training becomes smaller, as beam search allows for non-greedy local decisions that can potentially lead to a sequence with a higher overall probability, therefore is also tolerable to temporary mistakes.
\textcolor{black}{And even with the help of beam search, DSD loss can still bring improvement, which shows the effectiveness of DSD training in jumping out of the local optimum.}

\textcolor{black}{Since the neural machine translation model only penalizes the model with the top-$1$ prediction loss during training, the essence of DSD loss training is to optimize the loss in the opposite direction, so that the model does not trust the model's top-$1$ prediction too much, thereby improving the top-$1$ prediction in the inference stage. But in the inference stage, due to the adoption of beam search, top-$n$ prediction are considered, thus the advantage of DSD on top-$1$ is reduced. In order to verify our hypothesis, we conducted experiments on the Transformer-based NMT model optimized by ML+smoothing and cDSD under different beam sizes, respectively.}

\begin{table}[]
    \centering
    \caption{\textcolor{black}{Translation performance for different beam widths on English-German and English-French test sets.}} \label{tab:beam_size}
    \begin{tabular}{c|c|ccccc}
        \toprule
        \bf Model & \bf Dataset &  \bf B=1 & \bf B=3 & \bf B=5 & \bf B=25 & \bf B=100\\
        \midrule
        \multirow{2}{*}{Baseline} & En-De & 27.48 & 28.19 & 28.32 & 27.22 & 25.27 \\
        & En-Fr & 41.18 & 41.73 & 41.79 & 41.55 & 39.70 \\
        \midrule
        \multirow{2}{*}{+cDSD} & En-De & 28.00 & 28.46 & 28.64 & 27.87 & 26.11 \\
        & En-Fr & 41.79 & 41.92 & 42.00 & 41.91 & 40.45 \\
        \bottomrule
    \end{tabular}
\end{table}

\textcolor{black}{Table \ref{tab:beam_size} present the performance comparison. When the beam width is 1 (greedy search), we found that our cDSD training has the most obvious effect improvements compared to ML+smoothing training, which verifies our hypothesis in which cDSD can more effectively improve the prediction of top-$1$, thereby improving the final translation performance. When the beam size increases, because beam search has the ability to tolerate top-$1$ errors, the advantage of cDSD over ML+smoothing becomes smaller. Additionally, the translation performance degrades with large beam width due to the increasing beam width leads to sequences that are disproportionately based on early, very low probability tokens that are followed by a sequence of tokens with higher (conditional) probability, which is consistent with previous reports of similar performance degradation \cite{koehn-knowles-2017-six,ott2018analyzing,cohen2019empirical}.}

\subsection{Improvement over Back-translation}

\textcolor{black}{Back-translation is one of the most powerful enhancement strategies for machine translation. We also performed experiments on a stronger baseline enhanced with back-translation to demonstrate the effects of the DSD training rather than the variance of the model. We sampled 4.5M monolingual sentences from German news-crawl 2019 for back-translation. We first used the original WMT14 4.5M parallel corpus to train a backward German$\rightarrow$English translation model, and then based on the obtained backward translation model, the 4.5M monolingual was decoded and then combined with the parallel data for English$\rightarrow$German translation training. In the experiment, two settings: greedy search (B=1)and beam search (B=5) are adopted respectively (the respective beam width is also used in the backward decoding), and the final results are shown in Table \ref{tab:bt}.}

\begin{table}[]
    \centering
	\small
	\caption{\textcolor{black}{BLEU scores with models enhanced with back-translation on the English-German test set.}}\label{tab:bt}
    \begin{tabular}{c|ccc}
         \toprule
         & \bf B=1 & \bf B=5 & \bf $\Delta$ \\
         \midrule
         Baseline & 28.55 & 29.37 & 0.82 \\
         cDSD & 29.15 & 29.70 & 0.55 \\
         \bottomrule
    \end{tabular}
\end{table}

\textcolor{black}{After enhanced with back-translation, the performance of the baseline model has been improved significantly for both greedy search and beam search. On this strong baseline, cDSD has achieved similar improvements as on the original baseline. The performance difference between greedy search and beam search under cDSD training has become smaller, which shows that cDSD can improve the prediction of top-$1$, and further verify the source of improvements with cDSD training. In addition, the consistent improvement under a variety of experimental settings shows that the improvement of model performance does not come from the variance of training, but the better model robustness.}

\subsection{Comparison between DSD and Optimizer Switch Finetuning}

\textcolor{black}{In the use of DSD loss, we first use XENT+smoothing and Adam optimizer for fast convergence training, and then switch to DSD to train with the SGD optimizer. This optimizer switch finetuning also has the potential to make the model jump out of the local optimum. In order to show that the enhancement of DSD loss comes from the actual loss design rather than the optimizer switching finetuning, we introduce another baseline - OSF, in which the same steps are used to switch the Adam to SGD optimizer, but the original XENT+smoothing loss function is maintained.}

\begin{table}[!htbp]
	\centering
	\small
	\caption{\textcolor{black}{Difference of cDSD training and Optimizer Switch Finetuning (OSF) on English-German test set.}}\label{tab:switch}
	\begin{tabular}{cccc}
		\toprule
		\textbf{Model} & \textbf{Baseline} & \bf cDSD & \bf OSF \\
		\midrule
		\textbf{BLEU} & 28.32 & 28.64 & 28.35 \\
		\bottomrule
	\end{tabular}
\end{table}

\textcolor{black}{As shown in the comparison in Table \ref{tab:switch}, OSF has a very slight performance improvement over the baseline, but it is not significant, while the cDSD loss is relatively improved. Therefore, we can conclude that the source of DSD loss does not depend on the optimizer switch finetuning, but the better and robust convergence.}

\section{Related Work}

Standard NMT systems commonly adopt word-level cross entropy loss to learn model parameters; however, this type of ML learning has been shown to be a suboptimal method for sequence model training \cite{RanzatoCAZ16}. 
A number of recent works have sought different training strategies or improvements for the loss function. 
One of these approaches was by \cite{Bengio2015}, who proposed gently changing the training process from a fully-guided scheme that uses true previous tokens in prediction to a scheme that mostly uses previously generated tokens instead. 
Some other works focused on the study of sequence-level training algorithms. 
For instance, \cite{shen-EtAl} applied minimum risk training (MRT) in end-to-end NMT.  
\cite{wiseman-rush} introduced a sequence-level loss function in terms of errors made during the beam search. 
Another sequence-level training algorithm, proposed by \cite{RanzatoCAZ16}, directly optimized the evaluation metrics and was built on the REINFORCE algorithm. 
Similarly, with a reinforcement learning-style scheme, \cite{BahdanauBXGLPCB16} introduced a \emph{critic} network to predict the value of an output token; given the policy of an \emph{actor} network. This results in a training procedure that is much closer in practice to the test phase; and allows the model to directly optimize for a task-specific score such as BLEU. 
\cite{N18-1033} presented a comprehensive comparison of classical structured prediction losses for seq2seq models. 
Different from these works, we intend to optimize the training loss while allowing for easy implementation and not introducing more complexities. 

In terms of using symmetric KL divergence as loss, \cite{jiang-etal-2020-smart,aghajanyan2021better} improves ML training with additional symmetric KL divergence as a smoothing-inducing adversarial regularizer to achieve more robust fine-tuning purposes.
Different from both of the above motivations, in this work, we propose a general-purpose loss approach to cope with sequence-to-sequence tasks like NMT rather than providing an auxiliary distance measure. We give a new and effective loss for such broad range of tasks by integrating two symmetric KL-divergence terms with clear enough model training focus. Our suggested solution in this work is more general and more convenient for use.
\textcolor{black}{In terms of implementation, the loss form of \cite{zhang2019regularizing} is somewhat similar to ours. By introducing two Kullback-Leibler divergence regularization terms into the NMT training objective, a novel model regularization method for NMT training is proposed to improve the agreement between translations generated by left-to-right (L2R) and right-to-left (R2L) NMT decoders. Unlike their aims, our DSD loss is to improve the distribution of predicted tokens rather than improve the agreement problem between bidirectional decoding.}

\section{Conclusion}

This work proposes a general and balanced loss function for NMT training called \emph{dual skew divergence}. 
Adopting a hybrid training strategy with both cross entropy and DSD training, we empirically verify that switching to a DSD loss after the convergence of ML training gives an effect similar to that of simulated annealing and allows the model move to a better optimum. 
While the proposed DSD loss effectively enhances the RNN-based NMT, it suffers from unsatisfactorily balancing two symmetrical loss terms for deeper models like the Transformer, thus we further propose a controllable DSD and mitigate this issue. Our proposed DSD loss enhancement methods improve our diverse baselines, demonstrating a very general loss improvement.

\ifCLASSOPTIONcaptionsoff
  \newpage
\fi

\bibliographystyle{IEEEtran}
\bibliography{reference}

\end{document}